\documentclass[twoside,11pt]{article}

\usepackage{jmlr2e}
\usepackage{enumitem}
\usepackage[font=small,labelfont=bf]{caption}
\usepackage{subfigure}
\usepackage[utf8]{inputenc}

\jmlrheading{}{2020}{}{}{}{}{Joelle Pineau, Philippe Vincent-Lamarre, Koustuv Sinha, Vincent Larivi\`ere, Alina Beygelzimer, Florence d'Alch\'e-Buc, Emily Fox, Hugo Larochelle. Corresponding author: Joelle Pineau (jpineau@cs.mcgill.ca)}

\ShortHeadings{Improving Reproducibility in Machine Learning Research}{Pineau et al.}
\firstpageno{1}

\begin{document}

\title{\vspace{-2.0cm}Improving Reproducibility in Machine Learning Research \\ (\textit{\large A Report from the NeurIPS 2019 Reproducibility Program})}

\author{\name Joelle Pineau \email jpineau@cs.mcgill.ca \\
       \addr School of Computer Science, McGill University (Mila)\\
       Facebook AI Research \\
       CIFAR
       \AND
       \name Philippe Vincent-Lamarre \email{philvlam@gmail.com} \\
       \addr Ecole de biblioth\`economie et des sciences de l'information, \\ 
       Universit\'e de Montr\'eal
       \AND
       \name Koustuv Sinha \email{koustuv.sinha@mail.mcgill.ca} \\
       \addr School of Computer Science, McGill University (Mila)\\
       Facebook AI Research
       \AND
       \name Vincent Larivi\`ere \email{vincent.lariviere@umontreal.ca} \\
       \addr Ecole de biblioth\'economie et des sciences de l’information, \\
       Universit\'e de Montr\'eal
       \AND 
       \name Alina Beygelzimer \email{beygel@yahoo-inc.com} \\
       \addr Yahoo! Research
       \AND 
       \name Florence d'Alch\'e-Buc \email {florence.dalche@telecom-paris.fr} \\
       \addr T\'el\'ecom Paris, \\ 
       Institut Polytechnique de France
       \AND 
       \name Emily Fox \email {ebfox@cs.washington.edu} \\
       \addr University of Washington\\
       Apple
       \AND
       \name Hugo Larochelle \email {hugolarochelle@google.com} \\
       \addr Google \\
       CIFAR
       }


\maketitle

\begin{abstract}
One of the challenges in machine learning research is to ensure that presented and published results are sound and reliable. Reproducibility, that is obtaining similar results as presented in a paper or talk, using the same code and data (when available), is a necessary step to verify the reliability of research findings. Reproducibility is also an important step to promote open and accessible research, thereby allowing the scientific community to quickly integrate new findings and convert ideas to practice. Reproducibility also promotes the use of robust experimental workflows, which potentially reduce unintentional errors. In 2019, the Neural Information Processing Systems (NeurIPS) conference, the premier international conference for research in machine learning, introduced a reproducibility program, designed to improve the standards across the community for how we conduct, communicate, and evaluate machine learning research. The program contained three components: a code submission policy, a community-wide reproducibility challenge, and the inclusion of the Machine Learning Reproducibility checklist as part of the paper submission process. In this paper, we describe each of these components, how it was deployed, as well as what we were able to learn from this initiative. 

\end{abstract}


\section{Introduction}

At the very foundation of scientific inquiry is the process of specifying a hypothesis, running an experiment, analyzing the results, and drawing conclusions. Time and again, over the last several centuries, scientists have used this process to build our collective understanding of the natural world and the laws that govern it. However, for the findings to be valid and reliable, it is important that the experimental process be repeatable, and yield consistent results and conclusions. This is of course well-known, and to a large extent, the very foundation of the scientific process. Yet a 2016 survey in the journal Nature revealed that more than 70\% of researchers failed in their attempt to reproduce another researcher’s experiments, and over 50\% failed to reproduce one of their own experiments \citep{baker500ScientistsLift2016}.

In the area of computer science, while many of the findings from early years were derived from mathematics and theoretical analysis, in recent years, new knowledge is increasingly derived from practical experiments. Compared to other fields like biology, physics or sociology where experiments are made in the natural or social world, the reliability and reproducibility of experiments in computer science, where the experimental apparatus for the most part consists of a computer designed and built by humans, should be much easier to achieve. Yet in a surprisingly large number of instances, researchers have had difficulty reproducing the work of others \citep{henderson2018deep}.

Focusing more narrowly on machine learning research, where most often the experiment consists of training a model to learn to make predictions from observed data, the reasons for this gap are numerous and include:

\begin{itemize}
    \item Lack of access to the same training data / differences in data distribution;
    \item Misspecification or under-specification of the model or training procedure;
    \item Lack of availability of the code necessary to run the experiments, or errors in the code;
    \item Under-specification of the metrics used to report results;
    \item Improper use of statistics to analyze results, such as claiming significance without proper statistical testing or using the wrong statistic test;
    \item Selective reporting of results and ignoring the danger of adaptive overfitting;
    \item Over-claiming of the results, by drawing conclusions that go beyond the evidence presented (e.g. insufficient number of experiments, mismatch between hypothesis \& claim).
\end{itemize}

We spend significant time and energy (both of machines and humans), trying to overcome this gap. This is made worse by the bias in the field towards publishing positive results (rather than negative ones). Indeed, the evidence threshold for publishing a new positive finding is much lower than that for invalidating a previous finding. In the latter case, it may require several teams showing beyond the shadow of a doubt that a result is false for the research community to revise its opinion. Perhaps the most infamous instance of this is that of the false causal link between vaccines and autism. In short, we would argue that it is always more efficient to properly conduct the experiment and analysis in the first place.

In 2019, the Neural Information Processing Systems (NeurIPS) conference, the premier international conference for research in machine learning, introduced a reproducibility program, designed to improve the standards across the community for how we conduct, communicate, and evaluate machine learning research. The program contained three components: a code submission policy, a community-wide reproducibility challenge, and the inclusion of the Machine Learning Reproducibility checklist as part of the paper submission process.

In this paper, we describe each of these components, how it was deployed, as well as what we were able to learn from this exercise. The goal is to better understand how such an approach is implemented, how it is perceived by the community (including authors and reviewers), and how it impacts the quality of the scientific work and the reliability of the findings presented in the conference’s technical program. We hope that this work will inform and inspire renewed commitment towards better scientific methodology, not only in the machine learning research community, but in several other research fields.

\section{Background}

There is a growing interest in improving reproducibility across scientific disciplines.  A full review of such work is beyond the scope of this paper, but the common motivation is to ensure transparency of scientific findings, a faster and more reliable discovery process, and high confidence in our scientific knowledge.  In support of this direction, several biomedical journals agreed in 2014 on a set of principles and guidelines for reporting preclinical research in a way that ensured greater reproducibility \citep{mcnutt14science}.

There are challenges regarding reproducibility that appear to be unique (or at least more pronounced) in the field of ML compared to other disciplines. The first is an insufficient exploration of the variables (experimental conditions, hyperparameters) that might affect the conclusions of a study. In machine learning, a common goal for a model is to beat the top benchmarks scores. However, it is hard to assert if the aspect of a model claimed to have improved its performance is indeed the factor leading to the higher score. This limitation has been highlighted in a few studies reporting that new proposed methods are often not better than previous implementations when a more thorough search of hyper-parameters is performed \citep{lucic2018gans, melis2017state}, or even when using different random parameter initializations \citep{bouthillierUnreproducibleResearchReproducible, henderson2018deep}.

The second challenge refers to the proper documentation and reporting of the information necessary to reproduce the reported results \citep{gundersen2018state}. A recent report indicated that 63.5\% of the results in 255 manuscripts were successfully replicated \citep{raff2019step}. Strikingly, this study found that when the original authors provided assistance to the reproducers, 85\% of results were successfully reproduced, compared to 4\% when the authors didn’t respond. Although a selection bias could be at play (authors who knew their results would reproduce might have been more likely to provide assistance for the reproduction), this contrasts with large-scale replication studies in other disciplines that failed to observe similar improvement when the original authors of the study were involved \citep{klein2019many}. It therefore remains to be established if the field is having a reproduction problem similar to the other fields, or if it would be better described as a reporting problem.

Thirdly, as opposed to most scientific disciplines where uncertainty of the observed effects are routinely quantified, it appears like statistical analysis is seldom conducted in ML research \citep{fordeScientificMethodScience2019, henderson2018deep}.

\subsection{Defining Reproducibility}

\begin{figure}
    \centering
    \includegraphics[width=0.45\textwidth]{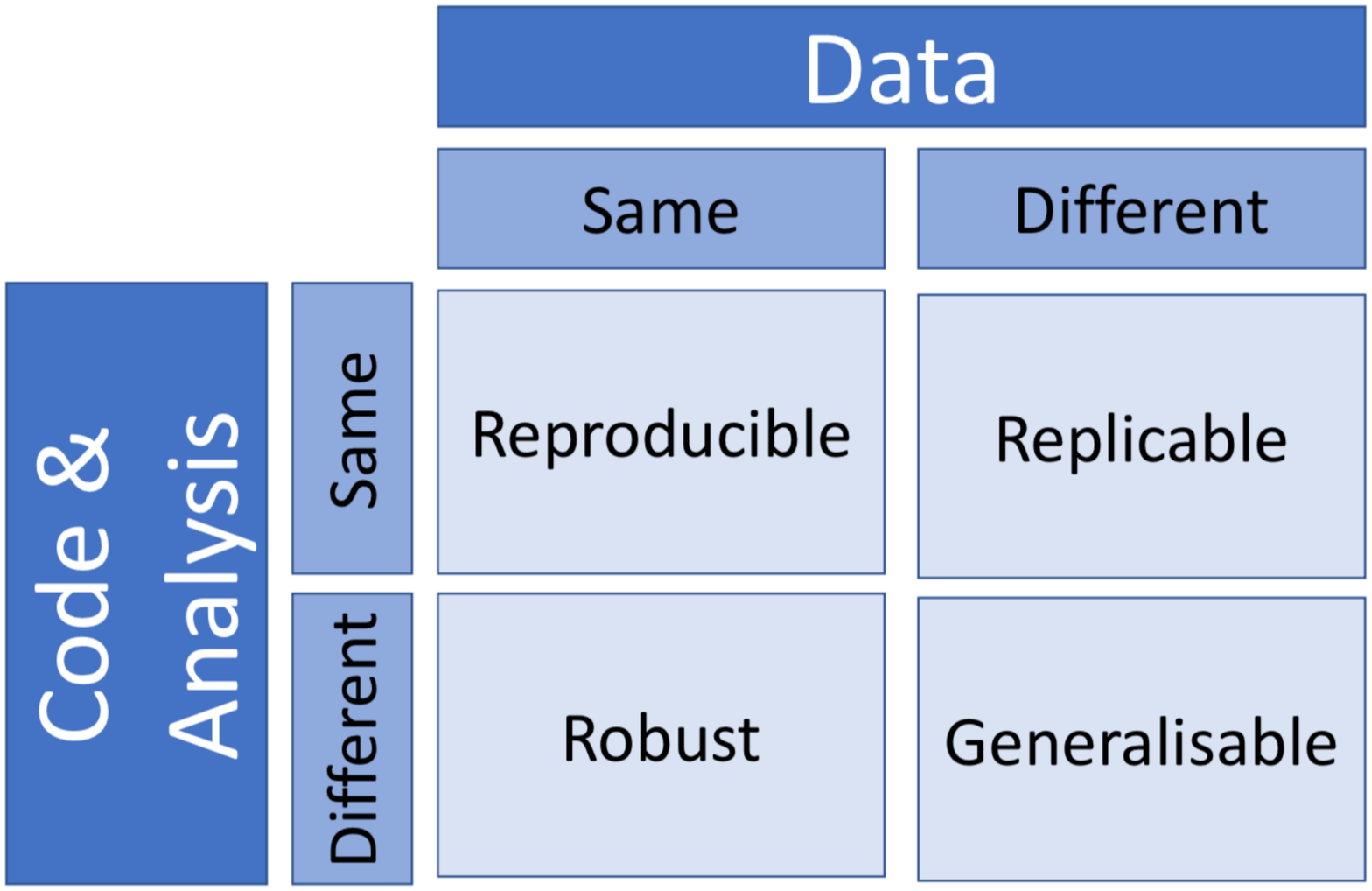}
    \caption{Reproducible Research. Adapted from: \href{https://github.com/WhitakerLab/ReproducibleResearch}{https://github.com/WhitakerLab/ReproducibleResearch}
    }
    \label{fig:define_rep}
\end{figure}

Before going any further, it is worth defining a few terms that have been used (sometimes interchangeably) to describe reproducibility \& related concepts. We adopt the terminology from Figure \ref{fig:define_rep}, where Reproducible work consists of re-doing an experiment using the same data and same analytical tools, whereas Replicable work considers different data (presumably sampled from similar distribution or method), Robust work assumes the same data but different analysis (such as reimplementation of the code, perhaps different computer architecture), and Generalisable work leads to the same conclusions despite considering different data and different analytical tools. For the purposes of our work, we focus primarily on the notion of Reproducibility as defined here, and assume that any modification in analytical tools (e.g. re-running experiments on a different computer) was small enough as to be negligible. A recent report by the National Academies of Sciences, Engineering, and Medicine, provides more in-depth discussion of these concepts, as well as several recommendations for improving reproducibility broadly across scientific fields \citep{national2019reproducibility}.

\subsection{The Open Science movement}

\textit{``Open Science is transparent and accessible knowledge that is shared and developed through collaborative networks”} \citep{vicente2018open}. In other words, Open science is a movement to conduct science in a more transparent way. This includes making code, data, scientific communications and any other research artifact publicly available and easily accessible over the long-term, thereby increasing the transparency of the research process and improving the reporting quality in scientific manuscripts \citep{sonnenburg2007need}. The implementation of Open science practices has been identified as a core factor that could improve the reproducibility of science \citep{munafo2017manifesto, gil:2016}. As such, the NeurIPS reproducibility program was designed to incorporate elements designed to encourage researchers to share the artefacts of their research (code, data), in addition to their manuscripts.

\subsection{Code submission policies}

It has become increasingly common in recent years to require the sharing of data and code, along with a paper, when computer experiments were used in the analysis. It is now standard expectation in the Nature research journals for authors to provide access to code and data to readers \citep{naturereport}. Similarly, the policy at the journal Science specifies that authors are expected to satisfy all reasonable requests for data, code or materials \citep{sciencejournals}. Within machine learning and AI conferences, the ability to include supplementary material has now been standard for several years, and many authors have used this to provide the data and/or code used to produce the paper. More recently, ICML 2019, the second largest international conference in machine learning has also rolled-out an explicit code submission policy \citep{icml19conf}.  During initial submission for double-blind reviewing, this can be uploaded as supplementary material.  For the final submission, it is best practice for data and code to be uploaded to a repository and given a DOI that can be cited.

\subsection{Reproducibility challenges}

Reproducibility tracks have been run prior in database systems conferences such as SIGMOD, as early as 2008 \citep{manolescu2008repeatability, 10.1145/2034863.2034873}, using the term ``repeatability" in lieu of our notion of reproducibility.  A notable recommendation was to focus on accepted papers (to spend effort on the work likely to have more impact).  Another useful recommendation was to provide  wiki page associated with each paper so the community could comment on it, post code, etc.  The organizers also found that the vast majority of author's whose work was included in a repeatability study found the process helpful.

The 2018 ICLR reproducibility challenge first introduced a dedicated platform to investigate reproducibility of papers for the Machine Learning community. The goal of this first iteration was to investigate reproducibility of empirical results submitted to the 2018 International Conference on Learning Representations \citep{ICLR2018Reproducibility2018}. The organizers chose ICLR for this challenge because the timing was right for course-based participants: most participants were drawn from graduate machine learning courses, where the challenge served as the final course project. The choice of ICLR was motivated by the fact that papers submitted to the conference were automatically made available publicly on OpenReview, including during the review period. This means anyone in the world could access the paper prior to selection, and could interact with the authors via the message board on OpenReview. This first challenge was followed a year later by the 2019 ICLR Reproducibility Challenge \citep{Pineau:2019}, and followed by 2019 NeurIPS Reproducibility Challenge \citep{SinhaRC:2020} in this edition.   Use of the OpenReview platform allowed a wiki-like interface to provide transparency into the replicability work and a conversation between authors and participants.

Several less formal activities, including hackathons, course projects, online blogs, open-source code packages, have participated in the effort to carry out re-implementation and replication of previous work and should be considered in the same spirit as the effort described here. 

\subsection{Checklists}

The Checklist Manifesto presents a highly compelling case for the use of checklists in safety-critical systems \citep{gawande2010checklist}. It documents how pre-flight checklists were introduced at Boeing Corporation as early as 1935 following the unfortunate crash of an airplane prototype. Checklists are similarly used in surgery rooms across the world to prevent oversights. Similarly, the WHO Surgical Safety Checklist, which is employed in surgery rooms across the world, has been shown to significantly reduce morbidity and mortality \citep{clay-williamsBackBasicsChecklists2015}.

In the case of scientific manuscripts, reporting checklists are meant to provide the minimal information that must be included in a manuscript, and are not necessarily exhaustive. 
The use of checklists in scientific research has been explored in a few instances.
Reporting guidelines in the form of checklists have been introduced for a wide range of study design in health research \citep{equatornetwork}, and the Transparency and Openness Promotion (TOP) guidelines have been adopted by multiple journals across disciplines \citep{nosek2015promoting}. There are now more than 400 checklists registered in the EQUATOR Network. CONSORT, one of the most popular guidelines used for randomized controlled trials was found to be effective and to improve the completeness of reporting for 22 checklist items \citep{turner2012consolidated}.

The ML checklist described below was significantly influenced by Nature’s Reporting Checklist for Life Sciences Articles \citep{naturecheck}. Other guidelines are under development outside of the ML community, namely for the application of AI tools in clinical trials \citep{liu2019extension} and health-care \citep{collinsReportingArtificialIntelligence2019}.

Concurrently with this work, a checklist was developed for AI publications that contain several of the same elements as we outline below, in terms of documenting data, code, methods and experiments \citep{gundersen18aimagazine}.  The checklist used at NeurIPS which we describe below is more oriented specifically towards machine learning experiments, with items related to test/validation/train splits, hyper-parameter ranges.

\subsection{Other considerations}

Beyond reproducibility, there are several other factors that affect how scientific research is conducted, communicated and evaluated. One of the best practices used in many venues, including NeurIPS, is that of double-blind reviewing. It is worth remembering that in 2014, the then program chairs Neil Lawrence and Corinna Cortes ran an interesting experiment, by assigning 10\% of submitted papers to be reviewed independently by two groups of reviewers (each lead by a different area chair). The results were surprising: overall the reviewers disagreed on 25.9\% of papers, but when tasked with reaching a 22.5\% acceptance rate, they disagreed on 57\% of the list of accepted papers. We raise this point for two reasons. First, to emphasize that the NeurIPS community has for many years already demonstrated an openness towards trying new approaches, as well as looking introspectively on the effectiveness of its processes. Second, to emphasize that there are several steps that come into play when a paper is written, and selected for publication at a high-profile international venue, and that a reproducibility program is only one aspect to consider when designing community standards to improve the quality of scientific practices.

\section{The NeurIPS 2019 code submission policy}

The NeurIPS 2019 code submission policy, as defined for all authors (see Appendix, Figure \ref{fig:code_submission_policy}), was drafted by the program chairs and officially approved by the NeurIPS board in winter 2019 (before the May 2019 paper submission deadline.)

\begin{figure}
	\centering
	\subfigure[]{
		\includegraphics[width=\linewidth]{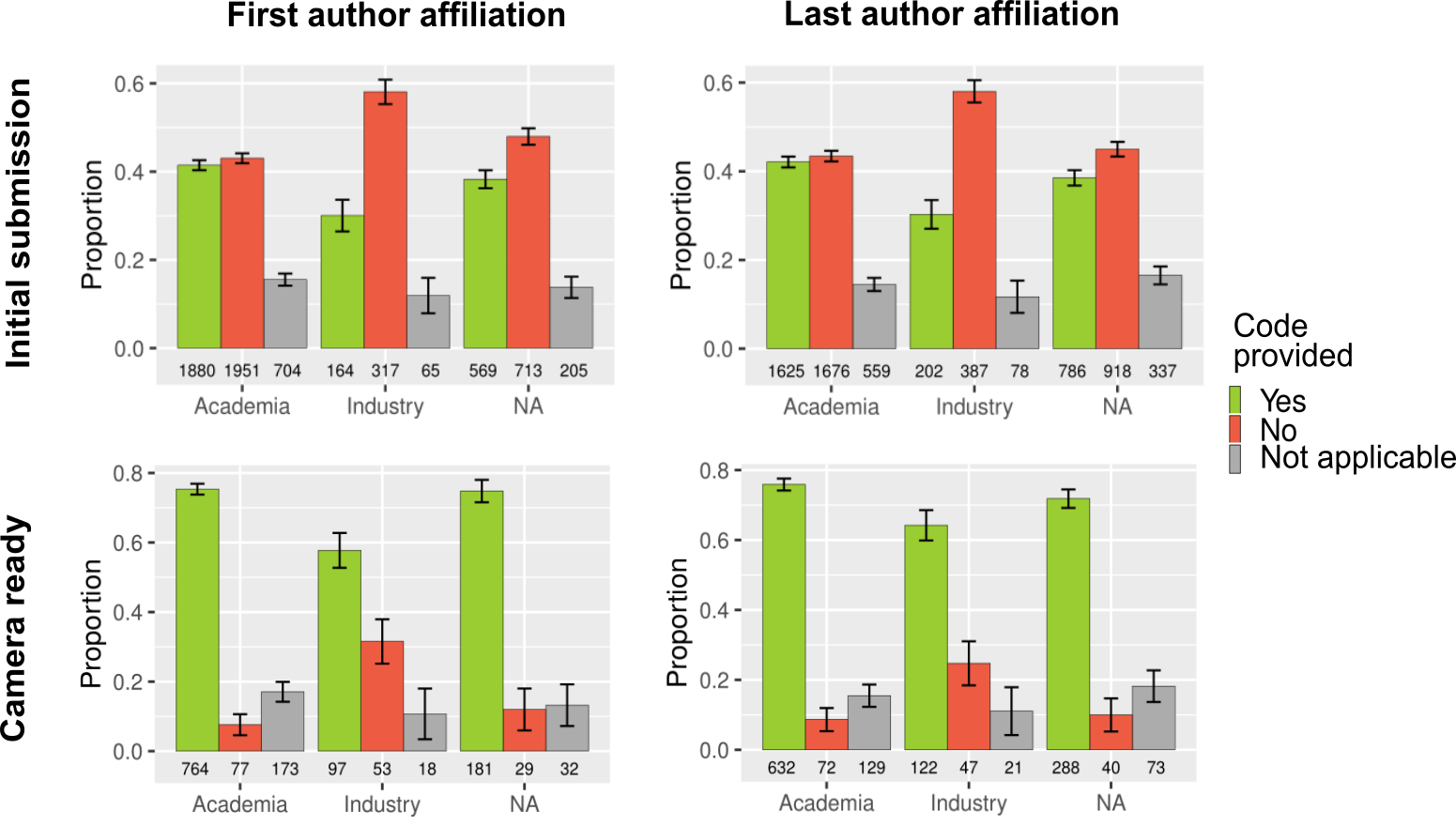}
		\label{fig:code_policy_a}}
	\subfigure[]{
		\centering
		\includegraphics[width=0.48\linewidth]{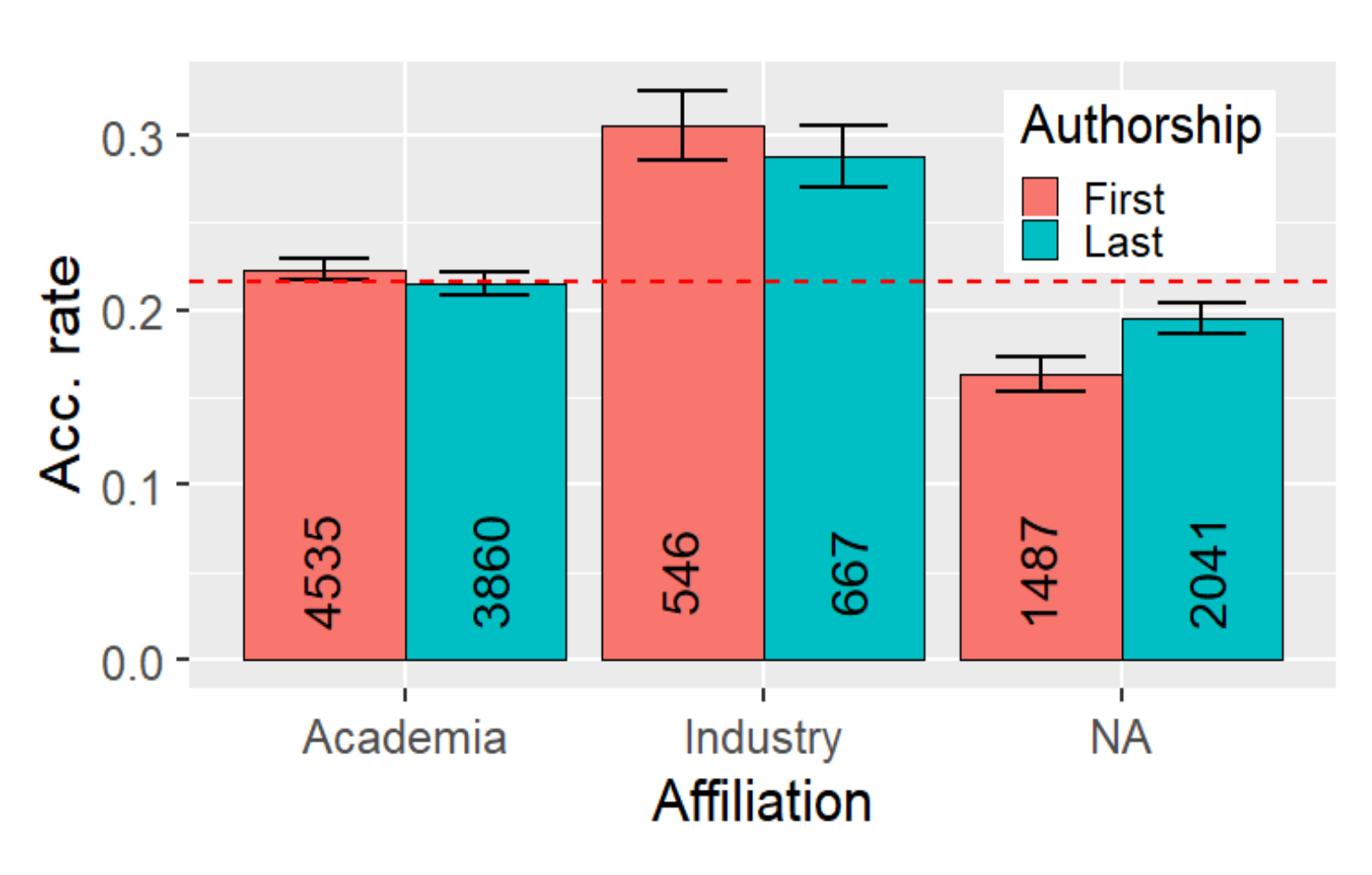}
		\label{fig:code_policy_b}}
	\caption{Effect of code submission policy. \ref{fig:code_policy_a} Link to code provided at initial submission and camera-ready, as a function of affiliation of the first and last authors. We observe for industry affiliated authors code is not provided in the initial submission, but later provided after camera ready. Overall, we observe authors from the academia are more prone to release the code of their papers. \ref{fig:code_policy_b}  Acceptance rate of submissions as a function of affiliation of the first and last authors. The red dashed line shows the acceptance rate for all submissions. We observe industry affiliated authors have higher chance of acceptance.}
\end{figure}

The most frequent objections we heard to having a code submission policy (at all) include:

\begin{itemize}
    \item  \textbf{Dataset confidentiality}: There are cases where the dataset cannot be released for legitimate privacy reasons. This arises often when looking at applications of ML, for example in healthcare or finance. One strategy to mitigate this limitation is to provide complementary empirical results on an open-source benchmark dataset, in addition to the results on the confidential data.
    \item \textbf{Proprietary software}: The software used to derive the result contains intellectual property, or is built on top of proprietary libraries. This is of particular concern to some researchers working in industry. Nonetheless, as shown in Figure \ref{fig:code_policy_a}, we see that many authors from industry were indeed able to submit code, and furthermore despite the policy, the acceptance rate for papers from authors in industry remained high (higher than authors from academia (Figure \ref{fig:code_policy_b})). By the camera-ready deadline, most submissions from the industry reported having submitted code (Figure \ref{fig:code_policy_a},\ref{fig:code_policy_b}). 
    \item \textbf{Computation infrastructure}: Even if data and code are provided, the experiments may require so much computation (time \& number of machines) that it is impractical for any reviewer, or in fact most researchers, to attempt reproducing the work. This is the case for work on training very large neural models, for example the AlphaGo game playing agent \citep{silver2016mastering} or the BERT language model \citep{devlinBERTPretrainingDeep2018}. Nonetheless it is worth noting that both these systems have been reproduced within months (if not weeks) of their release \citep{tian19opengo}.
    \item \textbf{Replication of mistakes}: Having a copy of the code used to produce the experimental results is not a guarantee that this code is correct, and there is significant value in reimplementing an algorithm directly from its description in a paper. This speaks more to the notion of Robustness defined above. It is indeed common that there are mistakes in code (as there may be in proofs for more theoretical papers). Nonetheless, the availability of the code (or proof) can be tremendously helpful to verify or re-implement the method. It is indeed much easier to verify a result (with the initial code or proof), then it is to produce from nothing (this is perhaps most poignantly illustrated by the longevity of the lack of proof for Fermat’s last theorem \citep{FermatLastTheorem2020}.)

\end{itemize}


\begin{figure}
	\centering
	\subfigure[]{
		\includegraphics[width=0.5\linewidth]{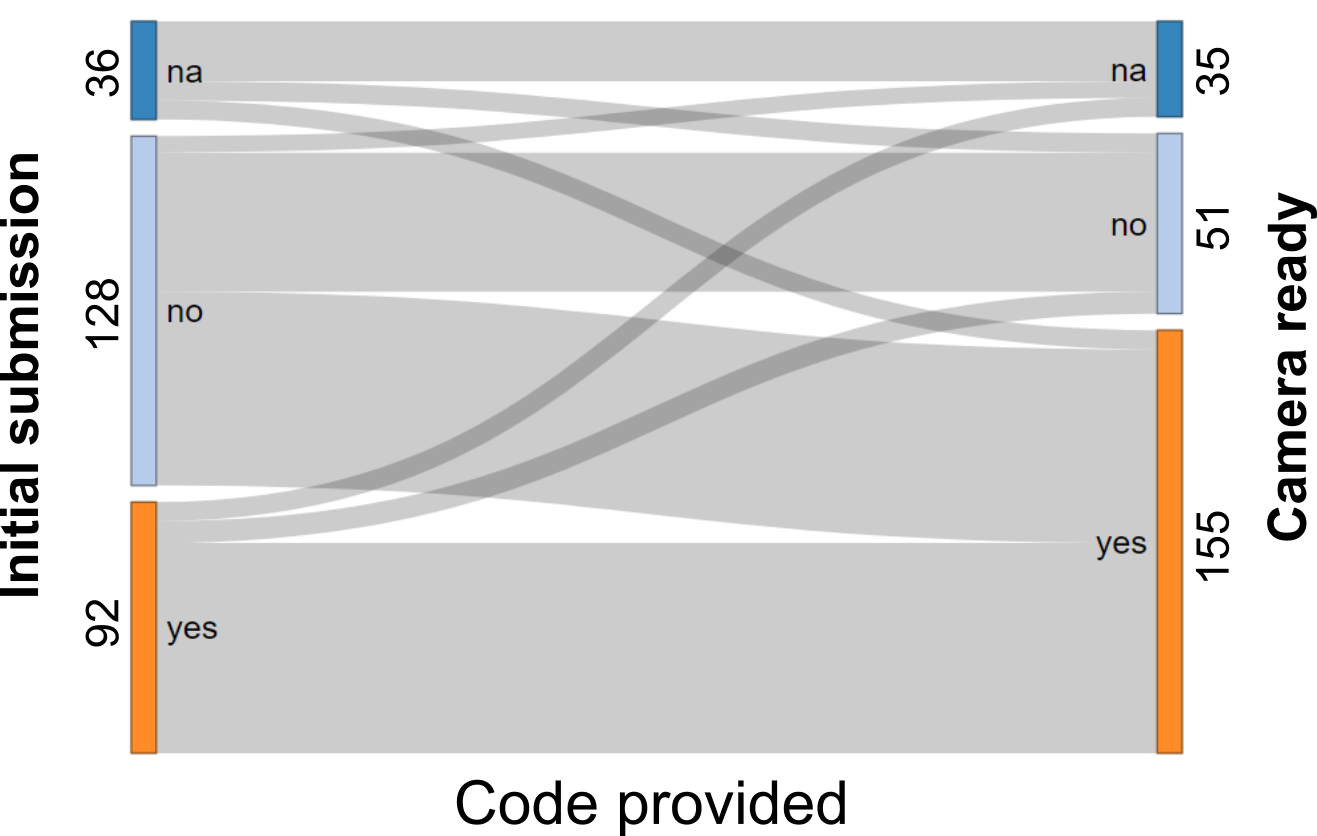}
		\label{fig:code_policy_c}}
	\subfigure[]{
		\centering
		\includegraphics[width=0.40\linewidth]{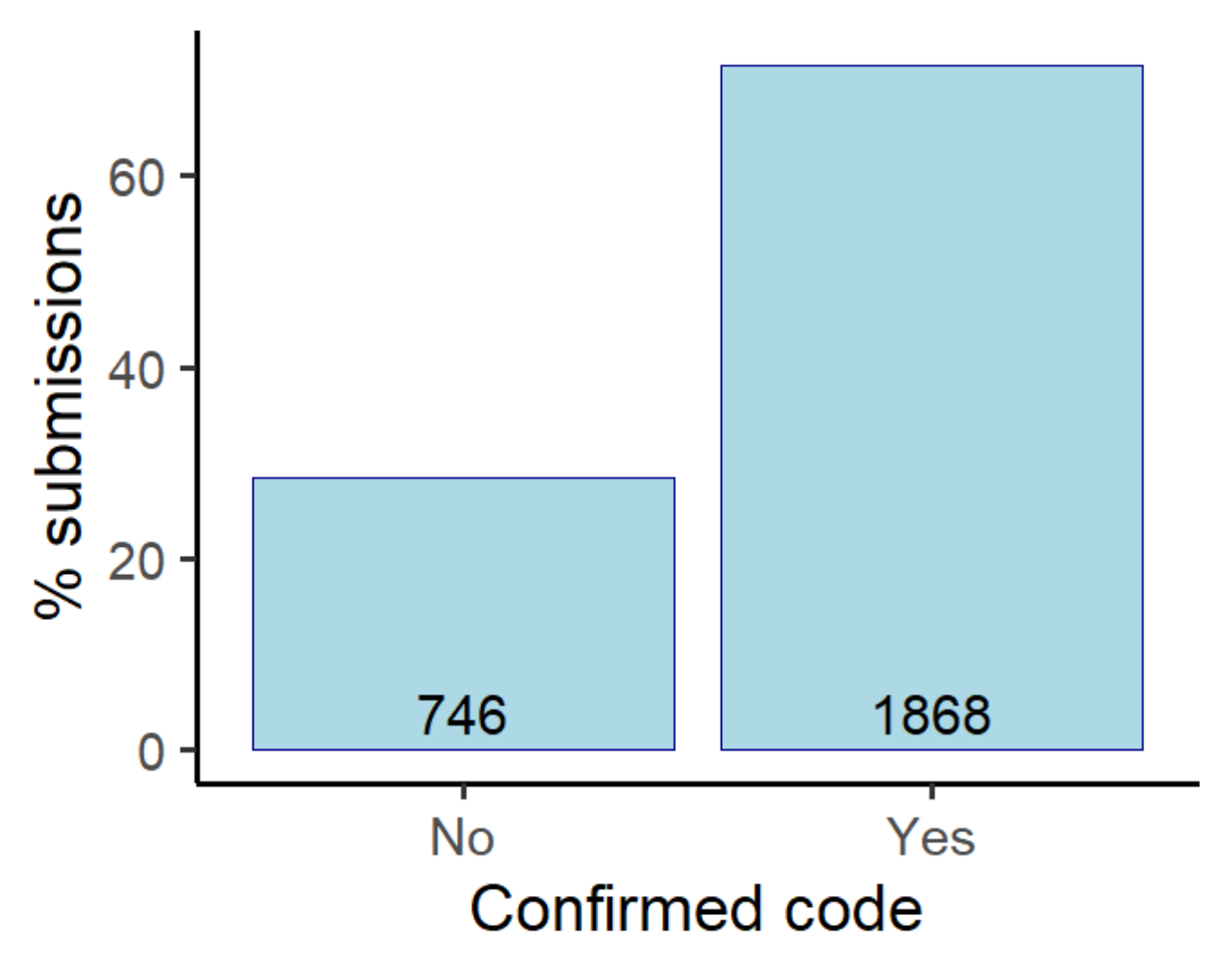}
		\label{fig:code_policy_d}}
	\caption{\ref{fig:code_policy_c} Diagram representing the transition of the code availability from initial submission to camera-ready only for submissions with an author from the industry (first or last). \ref{fig:code_policy_d}  Percentage of submissions reporting that they provided code on the checklist subsequently confirmed by the reviewers. }
\end{figure}


It is worth noting that the NeurIPS 2019 code submission policy leaves significant time \& flexibility, in particular it says that it: \textit{``\textbf{expects} code \textbf{only for accepted papers}, and only \textbf{by the camera-ready deadline}”}. So code submission is not mandatory, and the code is not expected to be used during the review process to decide on the soundness of the work. Reviewers were asked as a part of their assessment to report if code was provided along the manuscript at the initial submission stage. About 40\% of authors reported that they had provided code at this stage which was confirmed by the reviewers (if at least one reviewer indicated that the code was provided for each submission) for 71.5\% of those submissions (Figure \ref{fig:code_policy_d}). Note that authors are still able to provide code (or a link to code) as part of their initial submission. In Table \ref{tab:code_sub}, we provide a summary of code submission frequency for ICML 2019, as well as NeurIPS 2018 and 2019. We observe a growing trend towards more papers adding a link to code, even with only soft encouragement and no coercive measures.

\begin{table}[]
\centering
\resizebox{\textwidth}{!}{%
\begin{tabular}{|p{3cm}|p{2cm}|p{2cm}|p{3cm}|p{3cm}|p{5cm}|}
\hline
Conference & \# papers submitted & \% papers accepted & \% papers w/code at submission & \% papers w/code at camera-ready & Code submission policy \\ \hline
NeurIPS 2018 & 4856 & 20.8\% &  & \textless{}50\% & ``Authors may submit up to 100MB of supplementary material, such as proofs, derivations, data, or source code." \\ \hline
ICML 2019 & 3424 & 22.6\% & 36\% & 67\% & ``To foster reproducibility, we highly encourage authors to submit code. Reproducibility of results and easy availability of code will be taken into account in the decision-making process." \\ \hline
NeurIPS 2019 & 6743 & 21.1\% & 40\% & 74.4\% & “We expect (but not require) accompanying code to be submitted with accepted papers that contribute and present experiments with a new algorithm.” See Appendix, Fig. \ref{fig:code_submission_policy} \\ \hline
\end{tabular}%
}
\caption{Code submission frequency for recent ML conferences. Source for number of papers accepted and acceptance rates: \href{https://github.com/lixin4ever/Conference-Acceptance-Rate}{https://github.com/lixin4ever/Conference-Acceptance-Rate}. ICML 2019 numbers reproduced from \href{https://medium.com/@kamalika_19878/the-icml-2019-code-at-submit-time-experiment-f73872c23c55}{the ICML 2019 Code-at-Submit-Time Experiment}. 
}
\label{tab:code_sub}
\end{table}

While the value of having code extends long beyond the review period, it is useful, in those cases where code is available during the review process, to know how it is used and perceived by the reviewers. When surveying reviewers at the end of the review period, we found:

\begin{itemize}[label={}]
    \item \textit{Q. Was code provided (e.g. in the supplementary material)?} Yes: 5298
    \item \textit{If provided, did you look at the code?}  Yes: 2255
    \item \textit{If provided, was the code useful in guiding your review?} Yes: 1315
    \item \textit{If not provided, did you wish code had been available?} Yes: 3881
\end{itemize}

We were positively surprised by the number of reviewers willing to engage with this type of artefact during the review process. Furthermore, we found that the availability of code at submission (as indicated on the checklist) was positively associated with the reviewer score ($p<1e-08$).

\section{The NeurIPS 2019 Reproducibility Challenge}

The main goal of this challenge is to provide independent verification of the empirical claims in accepted NeurIPS papers, and to leave a public trace of the findings from this secondary analysis \citep{SinhaRC:2020}. The reproducibility challenge officially started on Oct.31 2019, right after the final paper submission deadline, so that participants could have the benefit of any code submission by authors. By this time, the authors’ identity was also known, allowing collaborative interaction between participants and authors. We used OpenReview \citep{neurips2019repchallenge} to enable communication between authors and challenge participants.


As shown in Table \ref{tab:rep_stats}, a total of 173 papers were claimed for reproduction. This is a 92\% increase since the last reproducibility challenge at ICLR 2019 \citep{Pineau:2019}. We had participants from 73 different institutions distributed around the world (see Appendix, Figure \ref{fig:participation}), including 63 universities and 10 industrial labs. Institutions with the most participants came from 3 continents and include McGill University (Canada), KTH (Sweden), Brown University (US) and IIT Roorkee (India). In those cases (and several others), high participation rate occurred when a professor at the university used this challenge as a final course project. 

\begin{table}[]
\centering
\resizebox{\textwidth}{!}{%
\begin{tabular}{|p{4cm}|p{2cm}|p{2cm}|p{2cm}|p{2cm}|p{2cm}|}
\hline
Conference & \# papers submitted & Acceptance rate & \# papers claimed & \# participating institutions & \# reports reviewed \\ \hline
ICLR 2018 & 981 & 32.0 & 123 & 31 & n/a \\ \hline
ICLR 2019 & 1591 & 31.4 & 90 & 35 & 26 \\ \hline
NeurIPS 2019 & 6743 & 21.1 & 173 & 73 & 84 \\ \hline
\end{tabular}%
}
\caption{Participation in the Reproducibility Challenge. Source for number of papers accepted and acceptance rates: \href{https://github.com/lixin4ever/Conference-Acceptance-Rate}{https://github.com/lixin4ever/Conference-Acceptance-Rate}
}
\label{tab:rep_stats}
\end{table}

All reports submitted to the challenge are available on OpenReview \footnote{\href{https://openreview.net/group?id=NeurIPS.cc/2019/Reproducibility_Challenge}{https://openreview.net/group?id=NeurIPS.cc/2019/Reproducibility\_Challenge}} for the community; in many cases with a link to the reimplementation code.  The goal of making these available is to two-fold:  first to give examples of reproducibility reports so that the practice becomes more widespread in the community, and second so that other researchers can benefit from the knowledge, and avoid the pitfalls that invariably come with reproducing another team’s work.
Most reports produced during the challenge offer a much more detailed \& nuanced account of their efforts, and the level of fidelity to which they could reproduce the methods, results \& claims of each paper.   Similarly, while some readers may be looking for a “reproducibility score”, we have not found that the findings of most reproducibility studies lend themselves to such a coarse summary.

Once submitted, all reproducibility reports underwent a review cycle (by reviewers of the NeurIPS conference), to select a small number of high-quality reports, which are published in the Sixth edition (Issue 2) \footnote{\href{https://rescience.github.io/read/\#issue-2-neurips-2019-reproducibility-challenge}{https://rescience.github.io/read/\#issue-2-neurips-2019-reproducibility-challenge}} of the journal ReScience \citep{resciencec, SinhaRC:2020}.  This provides a lasting archival record for this new type of research artefact.

\section{The NeurIPS 2019 ML reproducibility checklist}

The third component of the reproducibility program involved use of the Machine Learning reproducibility checklist (see Appendix, Figure \ref{fig:ml_checklist}). This checklist was first proposed in late 2018, at the NeurIPS conference, in response to findings of recurrent gaps in experimental methodology found in recent machine learning papers. An earlier version (v.1.1) was first deployed as a trial with submission of the final camera-ready version for NeurIPS 2018 papers (due in January 2019); this initial test allowed collection of feedback from authors and some minor modifications to the content of the checklist (mostly edited for clarity and reduced some redundant questions). The edited version 1.2 was then deployed during the NeurIPS 2019 review process, and authors were obliged to fill it both at the initial paper submission phase (May 2019), and at the final camera-ready phase (October 2019). This allowed us to analyze any change in answers, which presumably resulted from the review feedback (or authors’ own improvements of the work). The checklist was implemented on the CMT platform; each question included a multiple choice “Yes, No, not applicable”, and an (optional) open comment field.


\begin{figure}
    \centering
    \includegraphics[width=\textwidth]{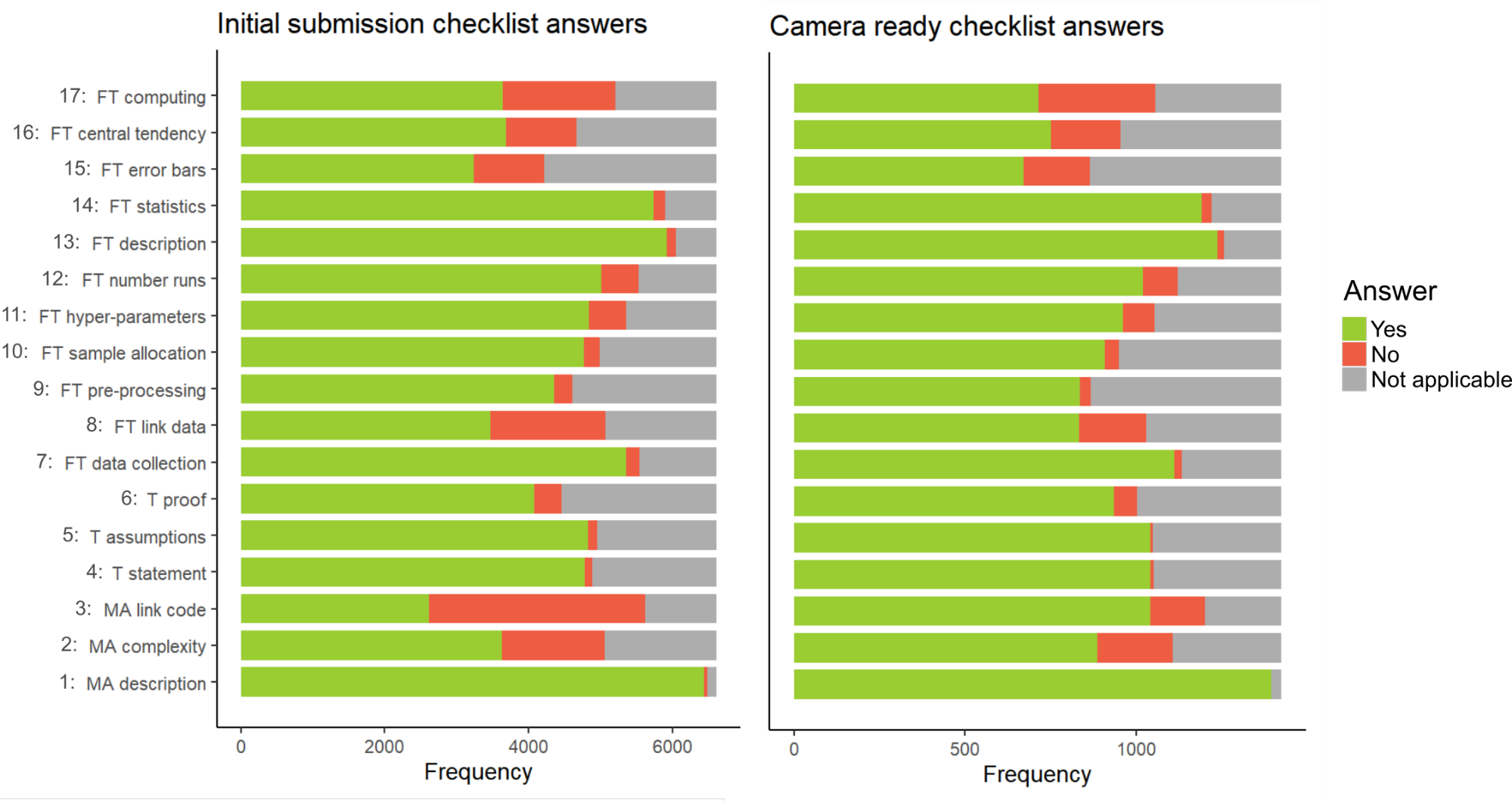}
    \caption{Author responses to all checklist questions for NeurIPS 2019 submitted papers. 
    }
    \label{fig:checklist}
\end{figure}

Figure \ref{fig:checklist} shows the initial answers provided for each submitted paper. It is reassuring to see that 97\% of submissions are said to contain Q\#. \textit{A clear description of the mathematical setting, algorithm, and/or model.} Since we expect all papers to contain this, the 3\% no/na answers might reflect margin of error in how authors interpreted the questions. Next, we notice that 89\% of submissions answered to the affirmative when asked Q\#. \textit{For all figures and tables that present empirical results, indicate if you include: A description of how experiments were run}. This is reasonably consistent with the fact that 9\% of NeurIPS 2019 submissions indicated “Theory” as their primary subject area, and thus may not contain empirical results.

One set of responses that raises interesting questions is the following trio:
\begin{itemize}[label={}]
    \item Q\#. \textit{A clear definition of the specific measure or statistics used to report results.}
    \item Q\#. \textit{Clearly defined error bars.}
    \item Q\#. \textit{A description of results with central tendency (e.g. mean) \& variation (e.g. stddev).}
\end{itemize}

In particular, it seems surprising to have 87\% of papers that see value in clearly defining the metrics and statistics used, yet 36\% of papers judge that error bars are not applicable to their results.

\begin{figure}
    \centering
    \includegraphics[width=\textwidth]{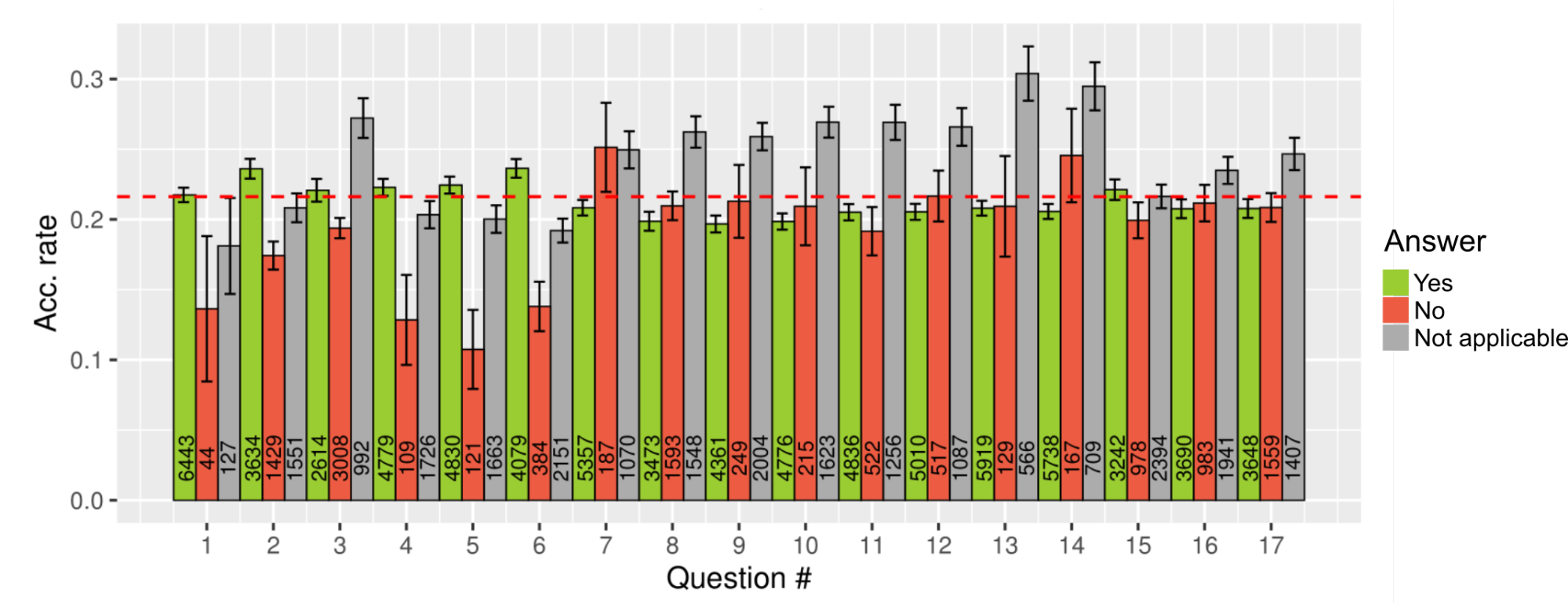}
    \caption{Acceptance rate per question. The x-axis corresponds to the question number on the checklist. The numbers within each bar show the number of submissions for each answer. See Fig. \ref{fig:checklist} (and in Appendix Fig. \ref{fig:ml_checklist}) for text corresponding to each Question \# (x-axis). The red dashed line shows the acceptance rate for all submissions. 
    }
    \label{fig:acceptance_rate}
\end{figure}

As shown in Figure \ref{fig:acceptance_rate}, many checklist answers appear to be associated with a higher acceptance rate when the answer is “yes”. However, it is too early to rule out potential covariates (e.g. paper’s topic, reviewer expectations, etc.) At this stage, it is encouraging that answering “no” to any of the questions is not associated with a higher acceptance rate. There seems to be a higher acceptance rate associated with “NA” responses on a subset of questions related to “Figures and tables”. Although it is still unclear at this stage why this effect is observed, it disappears when we only include manuscripts for which the reviewers indicated that the checklist was useful for the review. 

Finally, it is worth considering the reviewers’ point of view on the usefulness of the ML checklist to assess the soundness of the papers. When asked \textit{``Were the Reproducibility Checklist answers useful for evaluating the submission?”}, 34\% responded Yes.

\begin{figure}
    \centering
    \includegraphics[width=\textwidth]{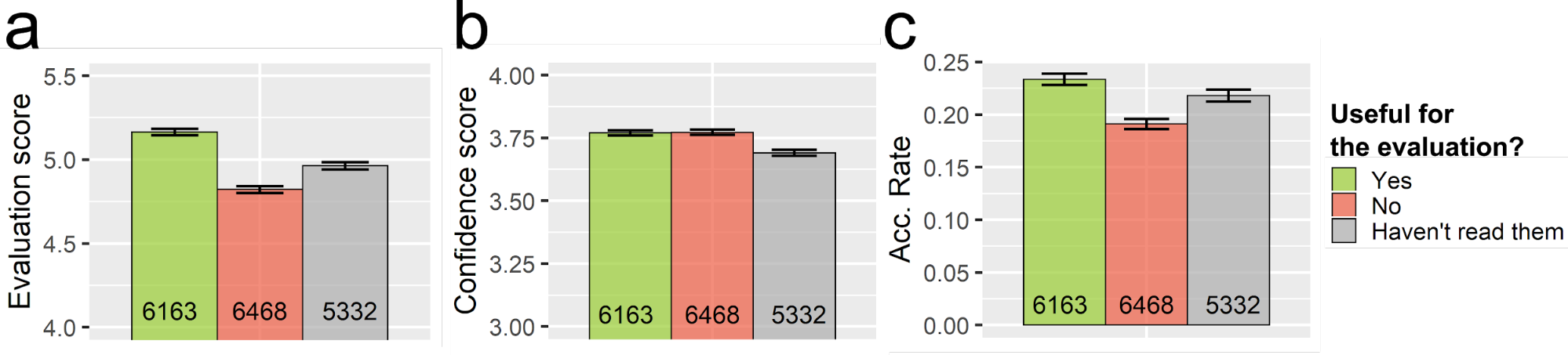}
    \caption{Perceived usefulness of the ML reproducibility checklist vs the review outcomes. (a) Effect on the paper score (scale 1-10). (b) Effect on the reviewer confidence score (scale of 1 to 5, where 1 is lowest). (c) Effect on the final accept/reject decision.
    }
    \label{fig:usefulness}
\end{figure}

We also note, as shown in Figure \ref{fig:usefulness}, that reviewers who found the checklist useful gave higher scores. And that those who found the checklist useful or not useful were more confident in their assessment than those who had not read the checklist. Finally, papers where the checklist was assessed as useful were more likely to be accepted.


\section{Discussion}

We presented a summary of the activities \& findings from the NeurIPS 2019 reproducibility program. Perhaps the best way to think of this effort is as a case study showing how three different mechanisms (code submission, reproducibility challenge, reproducibility checklist) can be incorporated into a conference program in an attempt to improve the quality of scientific contributions. At this stage, we do not have concluding evidence that these processes indeed have an impact on the quality of the work or of the papers that are submitted and published.

However we note several encouraging indicators:

\begin{itemize}
    \item  The number of submissions to NeurIPS increased by nearly 40\% this year, therefore we can assume the changes introduced did not result in a significant drop of interest by authors to submit their work to NeurIPS.
    \item The number of authors willingly submitting code is quickly increasing, from less than 50\% a year ago, to nearly 75\%. It seems a code submission policy based on voluntary participation is sufficient at this time. We are not necessarily aiming for 100\% compliance, as there are some cases where this may not be desirable (e.g. pre-processing script for confidential data).
    \item The number of reviewers indicating that they consulted the code, or wished to consult it is in the 1000’s, indicating that this is useful in the review process.
    \item The number of participants in the reproducibility challenge continues to increase, as does the number of reproducibility reports, and reviewers of reproducibility reports. This suggests that an increasing segment of the community is willing to participate voluntarily in secondary analysis of research results.
    \item One-third of reviewers found the checklist answers useful, furthermore reviewers who found the checklist useful gave higher scores to the paper, which suggests the checklist’s use is useful for both reviewers and authors.
\end{itemize}

The work leaves several questions open, which would require further investigation, and a careful study design to elucidate:

\begin{itemize}
    \item What is the long-term value (e.g. reproducibility, robustness, generalization, impact of follow-up work) of the code submitted?
    \item What is the effect of different incentive mechanisms (e.g. cash payment, conference registration, a point/badge system) on the participation rate \& quality of work in the reproducibility challenge?
    \item What is the benefit of using the checklist for authors?
    \item What is the accuracy of the ML checklist answers (for each question) when filled by authors?
    \item What is the measurable effect of the checklist on the quality of the final paper, e.g. in terms of soundness of results, clarity of writing?
    \item What is the measurable effect of the checklist on the review process, in terms of reliability (e.g. inter-rater agreement) and efficiency (e.g. need for response/rebuttal, discussion time)?
\end{itemize}

A related direction to explore is the development of tools and platforms that enhance reproducibility. Throughout this work we have focused on processes \& guidelines, but stayed away from prescribing any infrastructure or software tooling to support reproducibility. Many software tools, such as Docker\footnote{\url{https://www.docker.com/}}, ReproZip\footnote{\url{https://www.reprozip.org/}}, WholeTale\footnote{\url{https://wholetale.org/}} can encapsulate operating systems components, code, experimental variables and data files into a single package. Standardization of such tools would help sharing of information and improve ease of reproducibility.  A comparison of different tools is provided in \citep{isdahl19}.
We hope to see in the future some level of convergence on standadized tools for supporting reproducibility.

A few other CS conferences have developed reproducibility programs, and explored mechanisms beyond what we introduced at NeurIPS 2019.  For example, the ACM Multimedia conference, under the guidance of a Reproducibility Committee, has outlined specific reproducibility objectives, and included in 2021 a specific call for reproducibility papers\footnote{\url{https://project.inria.fr/acmmmreproducibility/}}. The ACM and its affiliated events has also introduced Badges\footnote{\url{https://www.acm.org/publications/policies/artifact-review-and-badging-current}} that can been attributed to papers to indicate when they meet pre-defined standards of Artifact availability, Reproducibility and Replication.   Beyond the CS community, other concrete suggestions have been provided to increase the trustworthiness in scientific findings \citep{jamison2019pnas}.

One additional aspect worth emphasizing is the fact that achieving reproducible results across a research community, whether NeurIPS or another, requires a significant cultural and organizational changes, not just a code submission policy or a checklist. The initiative described here is just one step in helping the community adopt better practices, in terms of conducting, communicating, and evaluating scientific research. The NeurIPS community is far from alone in looking at this problem. Several workshops have been held in recent years to discuss the issue as it pertains to machine learning and computer science \citep{ReproducibilitySIGCOMM17, ReproducibilityMachineLearning17, ReproducibilityMachineLearning18, ReproducibilityMachineLearning19}. Specific calls for reproducibility papers have been issued \citep{CallReproducibilityPapers}. An open-access peer-reviewed journal is dedicated to such papers \citep{resciencec}, which was used to publish select reports in ICLR 2019 Reproducibility Challenge \citep{Pineau:2019} and NeurIPS 2019 Reproducibility Challenge \citep{SinhaRC:2020}. 
And in the process, many labs are changing their practices to improve reproducibility of their own results. 

While this report focuses on the reproducibility program deployed for NeurIPS 2019, we expect many of the findings and recommendations to be more broadly applicable to other conferences and journals that represent machine learning research. Several other venues have already started defining code submission policies, though compliance varies. The ML checklist was crafted by consulting several other checklists, and could be adapted to other venues, as was done already for ICML 2020 and EMNLP 2020.  The 2020 version of the ML reproducibility challenge was also extended to cover papers accepted at 7 leading conferences (NeurIPS, ICML, ICLR, CVPR, EMNLP, CVPR and ECCV).  We do not foresee any major obstacles extending this to journal venues, where the longer review time and repeated interactions with the reviewers provide even more opportunity to meet high standards of reproducibility.


\acks{We thank the NeurIPS board and the NeurIPS 2019 general chair (Hanna Wallach) their unfailing support of this initiative. Without their courage and spirit of experimentation, none of this work would have been possible. We thank the many authors who submitted their work to NeurIPS 2019 and agreed to participate in this large experiment. We thank the program committee (reviewers, area chairs) of NeurIPS 2019 who not only incorporated the reproducibility checklist into their task flow, but also provided feedback about its usefulness. We thank Zhenyu (Sherry) Xue for preparing the data on NeurIPS papers \& reviews for the analysis presented here. We thank the OpenReview team (in particular Andrew McCallum, Pam Mandler, Melisa Bok, Michael Spector and Mohit Uniyal) who provided support to host the results of the reproducibility challenge. We thank CodeOcean (in particular Xu Fei) for providing free compute resources to reproducibility challenge participants. Thank you to Robert Stojnic and Yolanda Gil for valuable comments on an early version of the manuscript. Finally, we thank the several participants of the reproducibility challenge who dedicated time and effort to verify results that were not their own, to help strengthen our understanding of machine learning, and the types of problems we can solve today.}

\vskip 0.2in
\bibliography{main}

\newpage

\appendix


\begin{figure}[h]
    \centering
    \includegraphics[width=\textwidth]{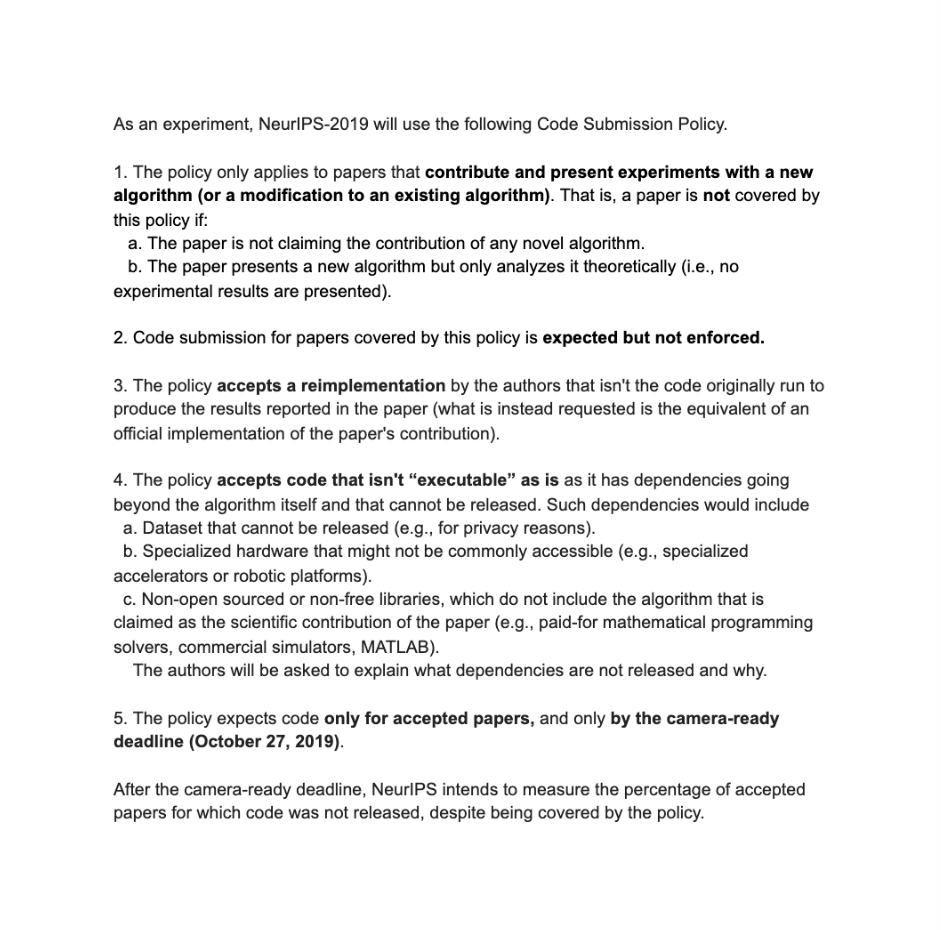}
    \caption{The NeurIPS 2019 code submission policy. Reproduced (with permission) from: [ADD URL
    }
    \label{fig:code_submission_policy}
\end{figure}

\begin{figure}[h]
    \centering
    \includegraphics[width=\textwidth]{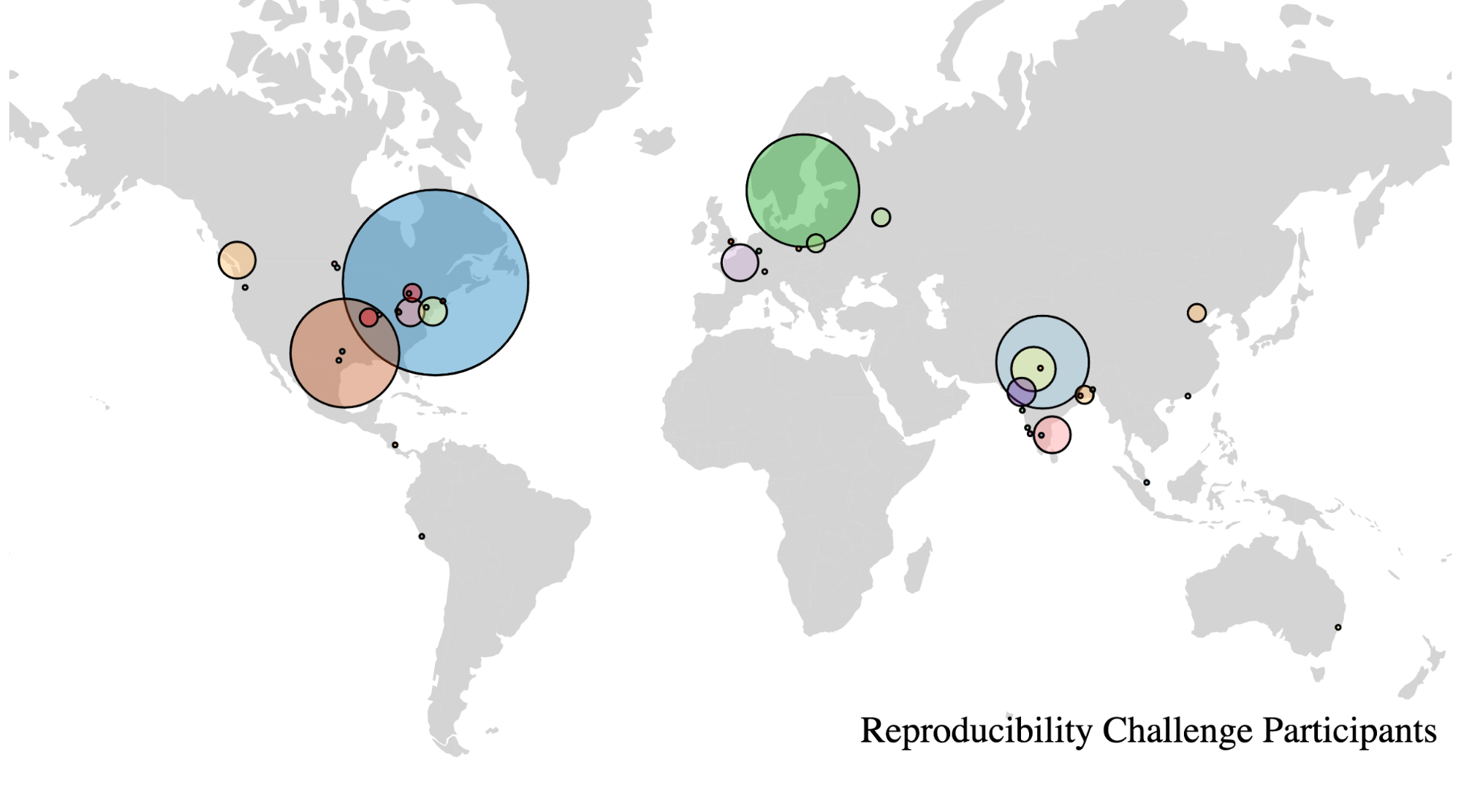}
    \caption{ NeurIPS 2019 Reproducibility Challenge Participants by geographical location.}
    \label{fig:participation}
\end{figure}

\begin{figure}[h]
    \centering
    \includegraphics[width=\textwidth]{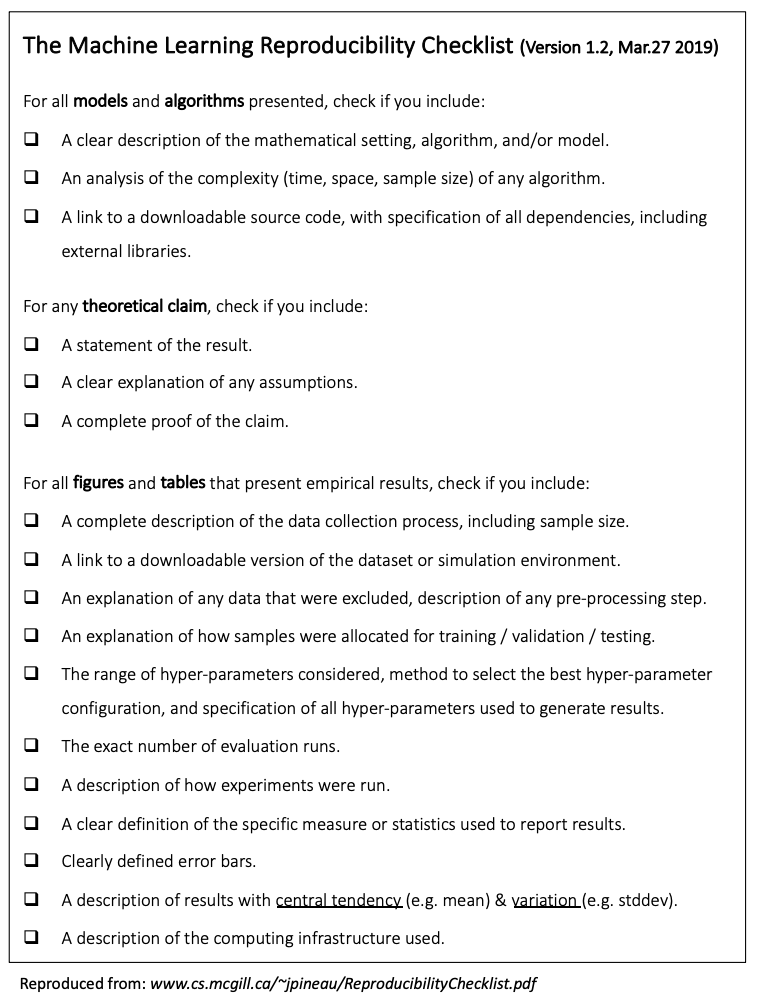}
    \caption{The Machine Learning Reproducibility Checklist, version 1.2, used during the NeurIPS 2019 review process.}
    \label{fig:ml_checklist}
\end{figure}

\end{document}